\PassOptionsToPackage{table,xcdraw}{xcolor}
\documentclass[runningheads]{llncs}

\usepackage{eccv}

\usepackage{eccvabbrv}

\usepackage{graphicx}
\usepackage{booktabs}

\usepackage[accsupp]{axessibility}

\usepackage{hyperref}
\usepackage{rotating}
\usepackage{pifont}
\usepackage{colortbl}
\usepackage{array}
\usepackage{multirow}
\usepackage{subcaption}
\usepackage{xcolor}
\usepackage{bibentry}
\usepackage{amsmath}
\usepackage{amssymb}
\usepackage{amsfonts}
\usepackage{booktabs}
\usepackage{makecell}
\usepackage{color}
\usepackage{soul}
\usepackage{threeparttable}
\usepackage{wrapfig}
\usepackage{subcaption}

\newcommand{\CheckmarkBold}{\ding{52}}

\newcommand{\myparagraph}[1]{\vspace{3pt}\noindent{\bf #1}}

\definecolor{barriercolor}{RGB}{255, 120, 50}
\definecolor{bicyclecolor}{RGB}{255, 192, 203}
\definecolor{buscolor}{RGB}{255, 255, 0}
\definecolor{carcolor}{RGB}{0, 150, 245}
\definecolor{constructcolor}{RGB}{0, 255, 255}
\definecolor{motorcolor}{RGB}{200, 180, 0}
\definecolor{pedestriancolor}{RGB}{255, 0, 0}
\definecolor{trafficcolor}{RGB}{255, 240, 150}
\definecolor{trailercolor}{RGB}{135, 60, 0}
\definecolor{truckcolor}{RGB}{160, 32, 240}
\definecolor{drivablecolor}{RGB}{0, 207, 191}
\definecolor{otherflatcolor}{RGB}{139, 137, 137}
\definecolor{sidewalkcolor}{RGB}{75, 0, 75}
\definecolor{terraincolor}{RGB}{150, 240, 80}
\definecolor{manmadecolor}{RGB}{222, 184, 135}
\definecolor{vegetationcolor}{RGB}{0, 175, 0}
\definecolor{otherscolor}{RGB}{0, 0, 0}

\usepackage{orcidlink}

\begin{document}

\title{Streaming Dense Voxel Representations\texorpdfstring{\\}{ }for 3D Occupancy Prediction}

\titlerunning{StreamOcc}

\author{Seokha Moon\inst{1,5}\ensuremath{^{\dagger}} \and
Janghyun Baek\inst{1} \and
Yujin Jeong\inst{2} \and
Daewon Chae\inst{3} \and \\
Giseop Kim\inst{4,5}\ensuremath{^{\ddagger}} \and
Jungbeom Lee\inst{1} \and
Jinkyu Kim\inst{1}\ensuremath{^{*}} \and
Sunwook Choi\inst{5}\ensuremath{^{*}}}

\authorrunning{S. Moon et al.}

\institute{{$^{1}$Korea University \quad
$^{2}$TU Darmstadt \& hessian.AI \quad
$^{3}$University of Michigan \quad
$^{4}$DGIST \quad
$^{5}$NAVER LABS}}

\makeatletter
\renewcommand{\@fnsymbol}[1]{\ensuremath{\ifcase#1\or *\or \dagger\or \ddagger\else\@ctrerr\fi}}
\makeatother

\maketitle
\begingroup
\renewcommand{\thefootnote}{\fnsymbol{footnote}}
\footnotetext[1]{Co-corresponding authors.}
\footnotetext[2]{Work done during an internship at NAVER LABS.\quad\footnotemark[3]~Work done at NAVER LABS.}
\endgroup
\vspace{-1.em}

\begin{abstract} 

In this paper, we explore dense voxel streaming for accurate and efficient 3D occupancy prediction. While dense voxel representations offer fine-grained spatial details and streaming paradigm enables efficient temporal processing, naively combining the two introduces key challenges: (i) warping-induced distortions caused by interpolation used for temporal alignment, and (ii) degraded dynamic object representations due to motion misalignment and detail loss in image-to-voxel projection. 
To address these, we propose \textbf{StreamOcc}, a novel framework that utilizes two aggregation strategies.
Specifically, it first refines propagated voxel features to reduce warping artifacts before temporal accumulation, and then selectively injects instance-level query features encoding dynamic-object semantics into the corresponding occupied voxel regions, preserving temporally consistent modeling while strengthening dynamic object representations.
Unlocking effective dense voxel streaming, StreamOcc achieves state-of-the-art performance on SurroundOcc-benchmark and Occ3D-nuScenes under real-time constraints, outperforming the prior best methods by \textbf{+1.3/2.5} and \textbf{+1.5/2.0 in (overall/dynamic object) mIoU}, respectively, while running at 83.3 ms per frame with only 2.8 GB of memory. The project page is available at \url{https://moonseokha.github.io/StreamOcc/}.


\end{abstract}

\begin{figure*}[t]
  \centering
    \includegraphics[width=1.\linewidth]{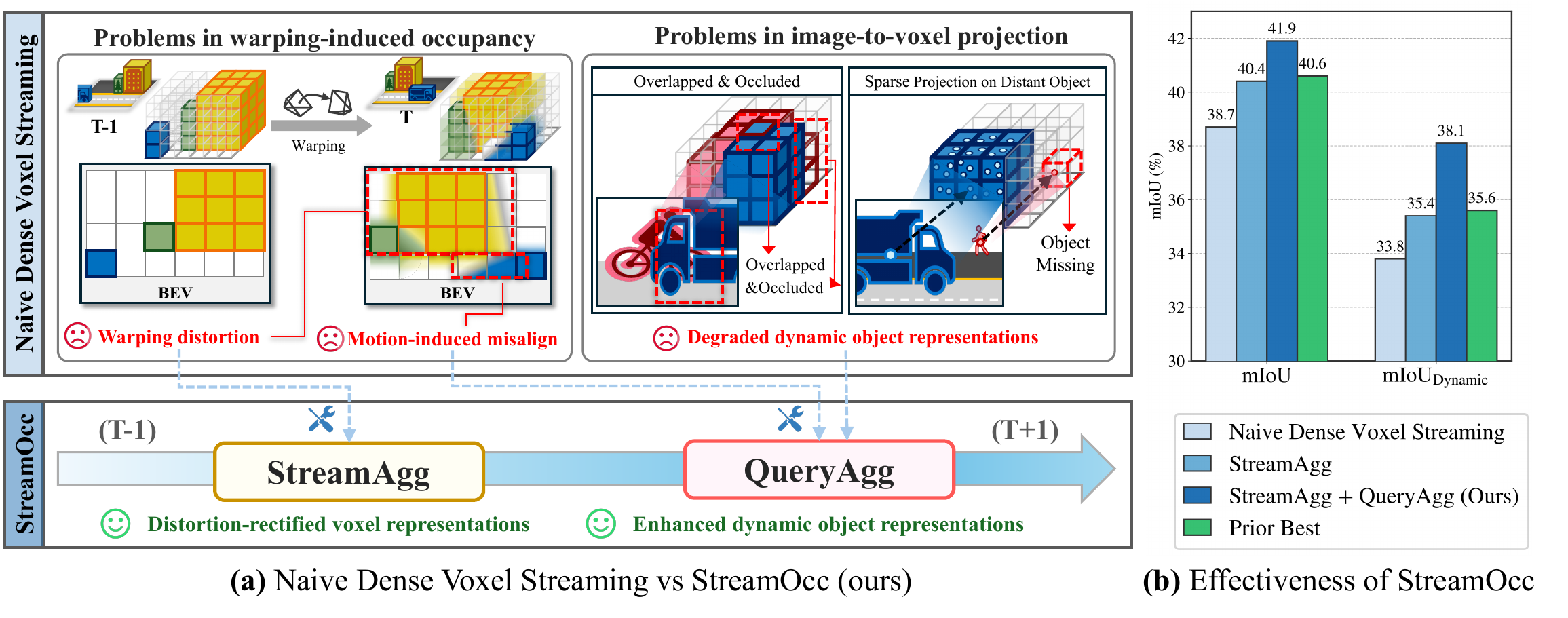}
    \vspace{-2em}
    \caption{\textbf{(a)} Challenges of naive dense voxel streaming and the proposed StreamOcc framework, consisting of StreamAgg and QueryAgg to address them. \textbf{(b)} 3D occupancy prediction results on Occ3D-nuScenes~\cite{occ3d}, showing consistent improvements with StreamOcc components over naive streaming and the prior best real-time method~\cite{alocc}.}
    \label{fig:teaser}
  \vspace{-2em}
\end{figure*}

\section{Introduction}
\label{sec:intro}

Vision-based 3D occupancy prediction has become a key perception task for autonomous driving, enabling dense and comprehensive scene understanding.
Specifically, it aims to classify each voxel in 3D space into semantic categories, including static objects (e.g., sidewalks, drivable surfaces), dynamic objects (e.g., vehicles, pedestrians), and free space~\cite{occ3d,surroundocc,occformer}.
This requires an accurate and fine-grained understanding of dense 3D spatial semantics.

Recent methods often leverage the multi-frame fusion mechanism (i.e., jointly processing current and several past frames) to incorporate temporal context, using either dense voxel representations~\cite{cotr,panoocc,bevdet4d,fbocc,bevformer,geocc,stcocc} or sparse representations~\cite{fastocc,gsdocc,sparseocc,opus,gaussianformer_v1,gaussianformer_v2} (e.g., z-axis pooled features, sparse voxels or queries, and Gaussians).
However, these methods suffer from an inherent accuracy–efficiency trade-off.
While dense voxel representations preserve fine-grained 3D spatial details, repeatedly processing dense historical features incurs high computational costs in terms of memory consumption (5–12GB) and inference latency (166–1,250ms), limiting practical deployment.
In contrast, compressed or sparse representations attempt to improve efficiency, but this often comes at the cost of spatial fidelity due to their limited representational capacity.

To avoid repeatedly processing multiple frames, the streaming paradigm offers an efficient alternative by recurrently updating propagated features with current-frame features. While this approach has demonstrated strong temporal modeling performance in sparse prediction tasks, such as 3D object and map detection~\cite{sparse4dv3,videobev,streammapnet}, its extension for 3D occupancy prediction remains non-trivial. Recent works have extended this streaming paradigm for 3D occupancy prediction, including GaussianWorld~\cite{gaussianworld} and ViewFormer~\cite{viewformer}. However, these approaches rely on sparse representation, such as Gaussian primitives or compressed representations, which inherently limit their capacity to accurately model dense 3D spatial semantics. 

Motivated by these observations, we explore a dense voxel streaming framework that integrates streaming-based temporal efficiency with voxel-based fine-grained 3D representation.
Yet, as illustrated in Fig.~\ref{fig:teaser} (a)-top, naively applying this integration struggles to capture fine-grained spatial details of the scene due to two key challenges:
\textbf{(i)} warping-induced distortions, which arise when aligning propagated voxel features from the previous timestep to the current ego-centric coordinate as grid values are resampled via interpolation.
\textbf{(ii)} degraded dynamic object features, which are caused by motion-induced misalignment and information loss that inherently occurs when projecting image features into voxel space (e.g., sparse projections of distant objects, coarse-grid instance merging, and occlusion-induced feature truncation).

In response to these challenges, our proposed method (\textbf{StreamOcc}) adopts two aggregation strategies to enable effective dense voxel streaming.
\textbf{First}, we design Rectified Voxel Streaming Aggregation (StreamAgg) to enable temporally consistent streaming of dense voxel features. It warps voxel features from the previous timestep and refines propagated voxel features to mitigate distortion through geometry-aware adaptive residual correction and align voxel semantics with the current frame before recurrent fusion.
\textbf{Second}, to improve dynamic object modeling, which is safety-critical in autonomous driving due to importance of interactions with dynamic agents, we introduce Query-guided Aggregation (QueryAgg).
In this module, instance queries encode dynamic object semantics extracted from image space and are selectively injected into the corresponding occupied voxel regions, compensating for the limitations of voxel-only accumulation.
Unlike prior strategies that re-aggregate image features across all voxels~\cite{cotr,geocc,bevformer}, our targeted aggregation focuses on dynamic objects, thereby enhancing their representations more effectively and efficiently.

Our contributions can be summarized as follows:
\begin{itemize}
    \item[$\bullet$] We introduce StreamOcc, the first framework that integrates a streaming paradigm with dense voxel representations for 3D occupancy prediction, achieving both high spatial fidelity and computational efficiency.
    \item [$ \bullet $] To mitigate warping-induced distortions and degraded dynamic object representations arising from streaming dense voxel features, we introduce Rectified Voxel Streaming Aggregation and Query-guided Aggregation; their effects are shown in Fig.~\ref{fig:teaser}(b).
    \item[$\bullet$] We demonstrate that StreamOcc achieves state-of-the-art performance on SurroundOcc-benchmark and  Occ3D-nuScenes under real-time constraints. Notably, it runs at 83.3 ms latency with only 2.8 GB memory, striking a strong balance between accuracy and efficiency.
\end{itemize}

\section{Related Work}
\label{sec:related_works}

\subsection{3D Occupancy Prediction}
3D occupancy prediction is a dense prediction task that assigns semantic labels to the entire scene-wise voxel space for comprehensive 3D scene understanding.
Early methods extended BEV features into 3D voxel space to construct voxel-based representations~\cite{bevdet, bevformer, bevstereo}, while subsequent approaches~\cite{surroundocc, occ3d, occformer, fbocc} further improved these voxel representations through multi-scale encoding and depth–semantic fusion.
However, using only a single frame limits visibility in occluded regions and results in sparse feature projection, leading to incomplete scene reconstructions.
To alleviate this limitation, multi-frame fusion methods~\cite{cotr, panoocc, geocc, bevdet4d, stcocc} jointly process consecutive frames to improve temporal consistency and scene completeness, achieving strong performance.
While multi-frame fusion improves temporal consistency and scene completeness, employing dense voxel representations in this setting introduces substantial computational overhead.
In pursuit of higher efficiency, recent methods adopt sparse or compressed representations.
TPVFormer~\cite{tpvformer} models 3D space using tri-plane features, while SparseOcc~\cite{sparseocc}, FastOcc~\cite{fastocc}, GSD-Occ~\cite{gsdocc}, and OPUS~\cite{opus} reduce computation through compressed features or sparse representations, including voxel- and query-based approaches. 
More recently, Gaussian-based methods~\cite{gaussianformer_v1, gaussianformer_v2, odg, gaussianworld} have emerged as efficiency-oriented alternatives. 
However, compressed and sparse representations often struggle to preserve fine-grained spatial fidelity or require a large number of primitives (e.g. Gaussians, query sets) to maintain semantic detail, leading to increased computational cost.

\subsection{Streaming in 3D Perception}
To avoid the computational cost of processing multiple frames, the streaming paradigm, which recurrently updates propagated features with current features, has emerged as an efficient alternative. 
This design has demonstrated strong temporal modeling performance in sparse prediction tasks, such as 3D object or map detection~\cite{sparse4dv3, streampetr, streamDSGN, videobev, streammapnet, onlinebev}. 
Motivated by these advantages, recent works have extended the streaming paradigm for 3D occupancy prediction using sparse representations, including GaussianWorld~\cite{gaussianworld}, which models the scene using Gaussian primitives, and ViewFormer~\cite{viewformer}, which adopts compressed representations (BEV feature).
However, unlike sparse prediction tasks, which model only a subset of spatial locations and do not require maintaining a fully dense representation, 3D occupancy prediction requires voxel-wise semantic modeling across the entire scene.
This fundamental difference makes sparse representations inherently insufficient for 3D occupancy prediction, as their limited representational capacity restricts accurate modeling of fine-grained 3D spatial semantics.
Building upon this observation, we propose StreamOcc, a framework that effectively integrates the streaming paradigm with dense voxel representations for accurate and efficient 3D occupancy prediction.

\section{Method}
\label{sec:method}

\begin{figure*}[t]
  \centering
    \includegraphics[width=.96\linewidth]{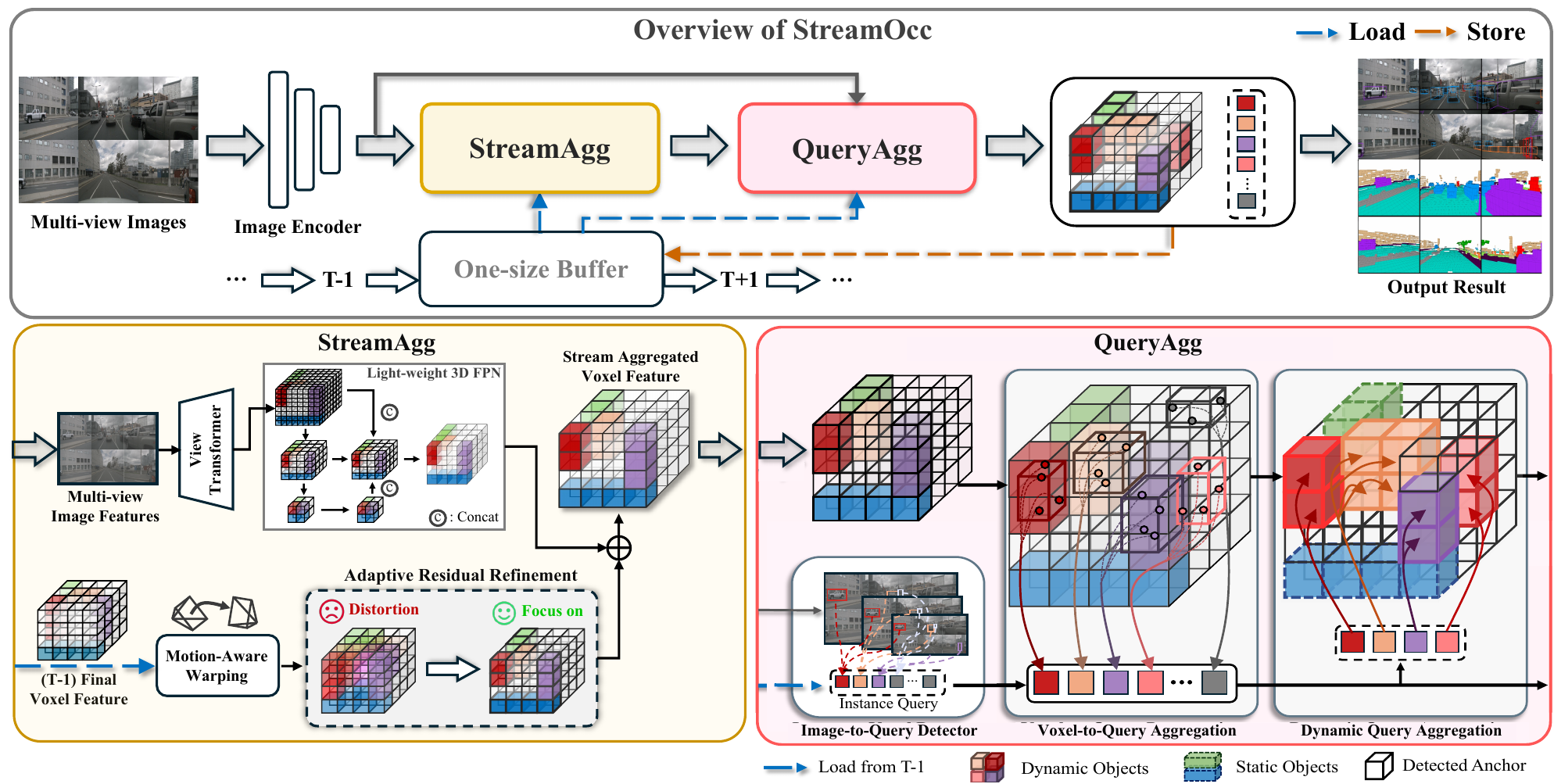}
  \vspace{-.5em}
  \caption{\textbf{Overview of StreamOcc}.
  StreamOcc predicts the 3D occupancy state of each voxel in a streaming manner via two-stage aggregation.
  First, voxel features are recurrently accumulated over time through Rectified Voxel Streaming Aggregation (StreamAgg), which mitigates warping-induced distortions during temporal propagation.  
  These temporally aggregated features are further refined using Query-guided Aggregation (QueryAgg), which leverages instance queries encoding fine-grained semantics to enhance dynamic object representations.
  }
  \vspace{-2.2em}
  \label{fig:overview}
\end{figure*}

We present a 3D occupancy prediction framework that effectively integrates a streaming paradigm with dense voxel representations. As illustrated in Fig.~\ref{fig:overview}, the framework consists of two main phases: Rectified Voxel Streaming Aggregation (StreamAgg; Sec.~\ref{sec:SVA}) and Query-guided Aggregation (QueryAgg; Sec.~\ref{sec:QGA}).

\subsection{StreamAgg: Rectified Voxel Streaming Aggregation}~\label{sec:SVA}
In this section, we present Rectified Voxel Streaming Aggregation (StreamAgg), which recurrently accumulates voxel features while mitigating warping-induced distortions to preserve temporal consistency. StreamAgg comprises three components: (i) 2D-to-3D feature extraction, (ii) motion-aware warping for temporal alignment, and (iii) adaptive residual correction to suppress the distortions.

\myparagraph{2D-to-3D View Transformation.}
Given a set of $N$ multi-view images $\{\mathbf{I}_i^{t}\}$ for $i \in \{1,2,\dots,N\}$ at the current timestep $t$, we extract multi-scale 2D features $\mathbf{F}_i$ using a ResNet~\cite{resnet} with FPN~\cite{fpn}.
Following prior works~\cite{cotr,gsdocc,bevdet4d}, we project the image features into a unified 3D voxel space, producing an initial voxel feature $\mathbf{V}^t_\text{init} \in \mathbb{R}^{C_\text{init} \times X \times Y \times Z}$, where $X$, $Y$, and $Z$ denote the voxel grid dimensions.
We then apply a lightweight 3D-FPN to aggregate multi-level voxel features and obtain the current downsampled voxel feature $\mathbf{V}_\text{down}^t \in \mathbb{R}^{C \times \frac{X}{2} \times \frac{Y}{2} \times \frac{Z}{2}}$.

\myparagraph{Motion-aware Voxel Feature Warping.}
In streaming settings, propagated voxel features $\mathbf{V}_{\text{final}}^{t-1} \in \mathbb{R}^{C \times \frac{X}{2} \times \frac{Y}{2} \times \frac{Z}{2}}$ from the previous timestep (which have recurrently accumulated temporal context) must be aligned to the current ego-centric coordinates.
Without such alignment, naive temporal accumulation leads to spatial inconsistencies, degrading dense voxel representations. To address this, we warp $\mathbf{V}_{\text{final}}^{t-1}$ according to the ego-motion between frames.

As the initial step of this warping process, we transform the spatial coordinates of the previous voxel grid $\mathbf{P}^e(t-1)$ into the current ego-centric frame:
\begin{equation}
    \mathbf{\bar{P}}^{e}(t) = \mathcal{T}_{g \to e }^{t} \cdot \mathcal{T}_{e \to g}^{t-1} \cdot \mathbf{P}^e(t-1),
\end{equation}
where $\mathcal{T}_{e \to g}^{t-1}$ maps coordinates from the past ego frame to the global frame, and $\mathcal{T}_{g \to e}^{t}$ maps them to the current ego-centric frame.
We then obtain the warped voxel feature via trilinear interpolation $\texttt{Interp}(\cdot)$ to resample past voxel features onto the current voxel grid:
\begin{equation}
    \mathbf{V}_{\text{warp}}^{t} = \texttt{Interp}(\mathbf{V}_{\text{final}}^{t-1}, \mathbf{\bar{P}}^{e}(t),\mathbf{P}^e(t-1)),
\end{equation}

\begin{figure}[t]
  \centering
    \includegraphics[width=.92\linewidth]{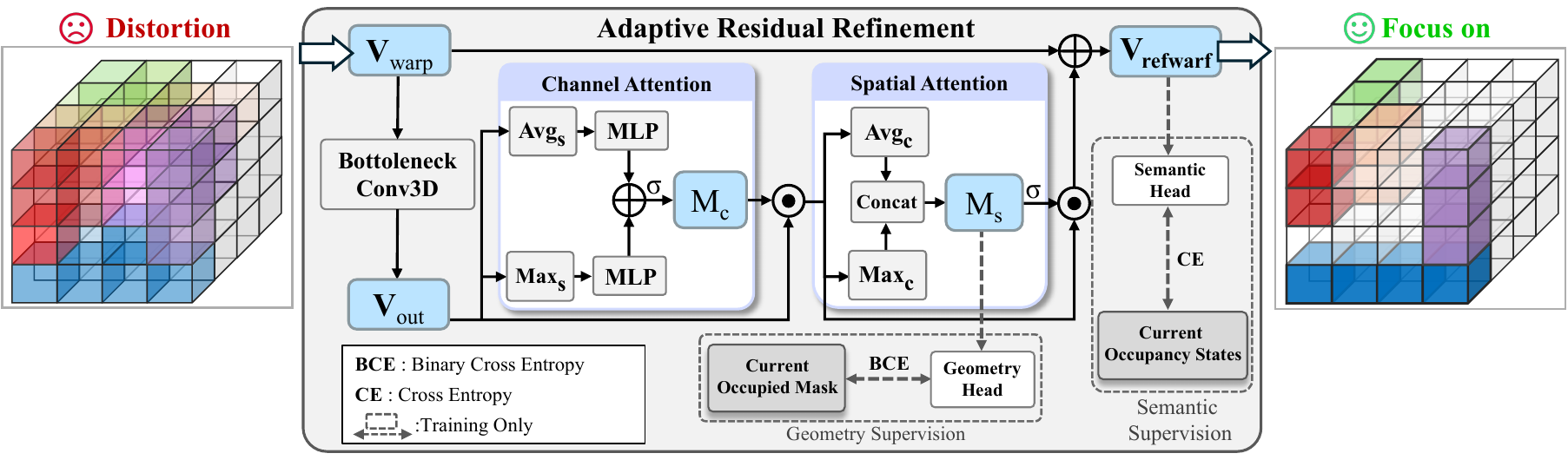}
    \vspace{-.5em}
  \caption{\textbf{Details of Adaptive Residual Refinement.} This module rectifies warping-induced distortions via adaptive residual correction that focuses on informative features guided by the geometric context. The refined warped voxel representation is supervised to achieve semantic alignment with the current scene state for consistent streaming. Geometry and Semantic heads are used only during training.}  
  \label{fig:refinenet}
  \vspace{-1.3em}
\end{figure}

\myparagraph{Adaptive Residual Refinement.}
The warping process propagates contextual cues from past observations but inevitably introduces distortions due to interpolation-based voxel features resampling.
These distortions often lead to erroneous values or blurred object boundaries, degrading feature quality and temporal consistency.
To mitigate these effects and preserve spatial consistency over time, we introduce an Adaptive Residual Refinement module (Fig.~\ref{fig:refinenet}).

Specifically, it adopts a bottleneck design~\cite{resnet} to efficiently process the warped feature $\mathbf{V}_{\text{warp}}^t$: it compresses the feature to $C/4$ channels, applies 3D convolutions, and expands it back to $C$ channels to obtain a corrective representation $\mathbf{V}_{\text{out}}^t$. Since warping artifacts are concentrated around object boundaries rather than uniformly distributed across the voxel space, the residual update should be selectively weighted. Accordingly, we extend the Convolutional Block Attention Module~\cite{cbam} to the 3D domain and introduce explicit supervision for adaptive residual correction and semantic refinement. Under this design, 3D channel and spatial attention are applied as follows:
\begin{equation}
\begin{aligned}
\mathbf{M}_c &= \sigma(\text{MLP}(\text{Avg}_{s}(\mathbf{V}_{\text{out}}^t)) + \text{MLP}(\text{Max}_{s}(\mathbf{V}_{\text{out}}^t))), \\
\mathbf{M}_s &= \text{Conv3D}([\text{Avg}_{c}(\mathbf{M}_c \odot \mathbf{V}_{\text{out}}^t); \text{Max}_{c}(\mathbf{M}_c \odot \mathbf{V}_{\text{out}}^t)]),
\end{aligned}
\label{eq:cbam3d}
\end{equation}
where $\sigma$ and $\odot$ denote the sigmoid function and element-wise multiplication, respectively, alongside spatial ($\text{Avg}_s$, $\text{Max}_s$) and channel ($\text{Avg}_c$, $\text{Max}_c$) pooling. 

The spatial attention feature $\mathbf{M}_s$ serves as a soft gating signal for the adaptive residual update.
To guide the gating mechanism to focus on meaningful features for the residual update, we supervise $\mathbf{M}_s$ with the ground-truth binary occupied mask during training \textbf{(Geometry Supervision)}.
This supervision enables $\mathbf{M}_s$ to distinguish occupied regions from free space, allowing geometry-aware adaptive weighting of the residual update.
Using these geometry-aware adaptive weights, we refine the warped voxel feature as follows:
\begin{equation}
\mathbf{V}_{\text{refwarp}}^t = \sigma(\mathbf{M}_s) \odot (\mathbf{M}_c \odot \mathbf{V}_{\text{out}}^t) + \mathbf{V}_{\text{warp}}^t.
\end{equation}

While geometry-guided gating encourages the residual update to focus on meaningful regions, it does not explicitly specify which semantic cues the corrective feature should learn to rectify warping-induced distortions in $\mathbf{V}_{\text{warp}}^t$.
To address this, we introduce \textbf{Semantic Supervision} to enforce semantic consistency by training $\mathbf{V}_{\text{refwarp}}^t$ to predict the semantic occupancy of the current frame.
This supervision guides the residual correction to encode semantically consistent features, suppressing warping artifacts and aligning the representation with the current scene state.

StreamAgg then outputs the aggregated voxel feature $\mathbf{V}_{\text{S.A}}^t$ by concatenating the refined warped feature with the current voxel feature and fusing them with a Conv1D block.
\begin{equation}
\mathbf{V}_{\text{S.A}}^t = \text{Conv}_{1D}\big([\mathbf{V}_{\text{refwarp}}^t;\mathbf{V}_{\text{down}}^t]\big).
\end{equation}

\subsection{QueryAgg: Query-guided Aggregation}
\label{sec:QGA}
Dynamic object features (e.g., vehicles, pedestrians) are particularly vulnerable to degradation due to motion misalignment and sparse image-to-voxel projection. (see Fig.~\ref{fig:visualization})
To address this limitation, we introduce Query-guided Aggregation (QueryAgg).
QueryAgg recurrently updates instance queries from multi-level image features $\mathbf{F}_i$ using Sparse4Dv3~\cite{sparse4dv3} in a streaming manner.
These queries are then enhanced with 3D geometric cues from voxel features and selectively aggregated into the corresponding occupied voxel regions, enabling accurate and robust dynamic object representation.

\myparagraph{Voxel-to-Query Aggregation.}  
Instance queries obtained from the Image-to-Query Detector~\cite{sparse4dv3} encode rich semantics and provide coarse localization.
However, due to depth ambiguity along the viewing ray, they often produce multiple false-positive queries across different depths (see Supplementary Material).
In contrast, voxel representations provide reliable 3D geometric information, but lack fine-grained semantic cues for dynamic objects.
This complementarity motivates the Voxel-to-Query Aggregation, which injects geometric information from $\mathbf{V}_{\text{S.A}}^t$ into instance queries to reduce depth ambiguity.

Specifically, we refine each query feature using deformable attention~\cite{deformdetr} to selectively sample informative regions in the voxel space. The refined query feature \( q_\text{ref}^{i} \) is updated as:
\begingroup
\setlength{\abovedisplayskip}{2pt}
\setlength{\belowdisplayskip}{2pt}
\begin{equation}
\label{eq:voxel_to_query}
\begin{array}{l}
q_{\text{ref}}^i = q_{\text{img}}^i + \sum_{h=1}^H W_h \bigg[ \sum_{o=1}^O \alpha_{iho} \cdot W'_h \cdot  \\
\quad \mathbf{V}_{\text{S.A}}^t\big(x_i + \Delta x_{iho}, y_i + \Delta y_{iho}, z_i + \Delta z_{iho}\big) \bigg],
\end{array}
\end{equation}
\endgroup
where \(q_{\text{img}}^i\) denotes the query feature produced by Image-to-Query Detector, and\( (x_i, y_i, z_i) \) denotes the 3D center of the associated object, and \( H \) and \( O \) are the numbers of attention heads and sampling points, respectively.
Each head predicts sampling offsets \((\Delta x_{iho}, \Delta y_{iho}, \Delta z_{iho})\) around the query center and aggregates voxel features from the StreamAgg representation \( \mathbf{V}_{\text{S.A}}^t \), using attention weights \( \alpha_{iho} \). 
The learnable matrices \( W_h \) and \( W'_h \) project query and sampled voxel features into a shared embedding space. 
This aggregation injects geometric cues into instance queries, reducing depth ambiguity and improving spatial localization.

\myparagraph{Dynamic Query Aggregation.} 
As mentioned above, voxel-only feature accumulation yields degraded representations for dynamic objects; therefore, we introduce Dynamic Query Aggregation (DQA), which selectively injects instance-level query features into voxel regions corresponding to dynamic objects.

Specifically, DQA first filters high-scoring instance queries (see the following subsection) that capture dynamic objects and maps them to the voxel regions occupied by the corresponding objects. Query-to-voxel attention is then applied to aggregate instance-level features into the voxel space as follows (in the following, we omit the timestep index $t$ for notational simplicity.):
\begin{gather}
\mathbf{q}^i = \mathbf{W}_Q \left( \mathbf{V}_{\text{S.A}}^i + \mathbf{p}_i \right), \quad 
\mathbf{k}^{ij} = \mathbf{W}_{K} q_{\text{ref}}^{ij}, \quad   
\mathbf{v}^{ij} = \mathbf{W}_{V} q_{\text{ref}}^{ij}, \\
\alpha^{i} = \text{softmax} \left( \frac{\mathbf{q}^i{}^\top}{\sqrt{d}} \cdot \left[ \mathbf{k}^{ij} \right]_{j \in \mathcal{N}^i} \right), \quad 
\mathbf{z}^i = \sum_{j \in \mathcal{N}^i} \alpha^{ij} \cdot \mathbf{v}^{ij}, \\
\mathbf{g}^i = \sigma \left( \mathbf{W}_{\text{gv}} \mathbf{V}_{\text{S.A}}^i + \mathbf{W}_{\text{gz}}\mathbf{z}^i \ \right),
\end{gather}
where $\mathcal{N}^i$ denotes the set of instance queries whose predicted bounding boxes overlap with the $i$-th voxel grid cell, and $q_{\text{ref}}^{ij}$ represents the feature of the $j$-th query associated with voxel $i$. Here, $\mathbf{q}^i$, $\mathbf{k}^{ij}$, and $\mathbf{v}^{ij}$ are obtained via linear projections $\mathbf{W}_Q$, $\mathbf{W}_K$, and $\mathbf{W}_V$, respectively, and $\mathbf{p}_i$ denotes positional embedding. Attention vector $\alpha^{i}$ is computed over $\mathcal{N}^i$, and $\alpha^{ij}$ denotes its $j$-th element, which weights each query feature when forming the aggregated feature $\mathbf{z}^i$.

For voxels with no overlapping queries ($\mathcal{N}^i = \emptyset$), the features remain unchanged; for voxels with overlapping queries, the gating vector $\mathbf{g}^i$ determines how much of the aggregated instance feature $\mathbf{z}^i$ should be injected into voxel $i$:
\begin{equation}
\mathbf{V}_{\text{DQA}}^i =
\begin{cases}
\mathbf{V}_{\text{S.A}}^i, &  \text{if } \mathcal{N}^i = \emptyset, \\
\mathbf{V}_{\text{S.A}}^i + \mathbf{g}^i \odot \mathbf{z}^i, & \text{otherwise}.
\end{cases}
\end{equation}
This selective update allows DQA to enhance dynamic-object voxels while preserving stable representations in static or empty regions. 
The updated voxel features $\mathbf{V}_{\text{DQA}}$ are further processed by a feed-forward network with normalization and residual connections, producing the final voxel representation $\mathbf{V}_{\text{final}}$, which is fed into the occupancy head for 3D occupancy prediction.

\myparagraph{Query Selection for DQA.}
Reliable query selection is essential for effective Dynamic Query Aggregation, as naive selection based solely on IoU may introduce shortcut learning.
In particular, IoU-only selection can map queries that happen to be close to the ground truth into voxel features regardless of detection quality, leading to implicit ground-truth leakage and degraded inference performance.
Therefore, we design a training-time query selection strategy that combines confidence scores (0.3) with either IoU or geometric constraints to maintain generalization.
The selection is scale-aware: we use an IoU-based criterion for large objects, while adopting a geometry-based criterion for small objects where IoU becomes unreliable, by enforcing consistency between predicted and ground-truth box geometry (e.g., center and size deviations); detailed thresholds and formulations are provided in the Supplementary Material.

During inference, the filtering strategy is simplified by selecting only instance queries with a confidence score above 0.3, consistent with the detection threshold.

\subsection{Decoder for Occupancy Prediction}
\label{sec:decoder}
To predict the high-resolution 3D voxel occupancy, we upsample the final voxel feature \( \mathbf{V}_{\text{final}} \) to obtain \( \mathbf{V}_{\text{up}} \in \mathbb{R}^{C \times X \times Y \times Z} \), which is processed by a lightweight MLP-based decoder used in training and inference.
To further enhance semantic representation learning without modifying the voxel features, we incorporate an Auxiliary Mask Decoder inspired by Co-DETR~\cite{codetr}, adopting the multi-head transformer-based mask prediction design from the Semantic-aware Group Decoder in COTR~\cite{cotr}.
During training, a group-wise one-to-many assignment associates each ground-truth mask with multiple queries, providing richer supervision.
As this auxiliary decoder is used solely for feature learning, it is omitted at inference time to avoid unnecessary computation and maintain efficiency.

\section{Experiments}
\label{sec:experiment}
\subsection{Datasets}
We evaluate our method on two benchmark datasets for 3D occupancy prediction: Occ3D-nuScenes~\cite{occ3d} and SurroundOcc dataset~\cite{surroundocc}. 
Both datasets represent the scene as a voxel grid of size $200 \times 200 \times 16$, where each voxel is labeled as either \textit{occupied} or \textit{empty}, and occupied voxels are further categorized into 16 semantic classes such as \texttt{car} and \texttt{pedestrian}. 
In \textbf{Occ3D-nuScenes,}
the voxel grid is defined in the ego coordinate, spanning $[-40\text{m}, 40\text{m}]$ along both $X$ and $Y$, and $[-1\text{m}, 5.4\text{m}]$ along $Z$, discretized at a resolution of $0.4\text{m}$. 
Occupied voxels are assigned 17 semantic classes (16 known categories + \texttt{others}).
In \textbf{SurroundOcc,}
the voxel grid is defined in the LiDAR coordinate, covering $[-50\text{m}, 50\text{m}]$ along $X$ and $Y$, and $[-5\text{m}, 3\text{m}]$ along $Z$, with a voxel resolution of $0.5\text{m}$. 

\subsection{Implementation and Evaluation Details}
Following standard practice~\cite{cotr,gsdocc,bevdet4d,alocc}, we use ResNet-50~\cite{resnet} as the image backbone and resize input images to $256 \times 704$. 
The initial voxel feature $\mathbf{V}_{\text{init}}$ is represented as a $200 \times 200 \times 16$ voxel grid with 64 channels. 
We use 900 instance queries for processing dynamic objects, and the auxiliary mask decoder consists of 6 heads.
We train with AdamW~\cite{adamw} for 24 epochs on Occ3D-nuScenes and 20 epochs on SurroundOcc, without CBGS~\cite{cbgs}, using a batch size of 8, gradient clipping, an initial learning rate of $2 \times 10^{-4}$, and a warmup ratio of $1/3$ for the first 200 iterations.
Inference latency and GPU memory are measured on a single NVIDIA A100 or RTX 4090 GPU. Further, we evaluate occupancy prediction using mean IoU (mIoU and mIoU$_{D}$ over semantic classes and dynamic objects), and additionally report class-agnostic IoU to assess overall 3D geometry reconstruction by treating all occupied voxels as foreground. The Image-to-Query detector used in QueryAgg is evaluated using nuScenes Detection Score (NDS)~\cite{nuscenes} and mean Average Precision (mAP).

\begin{table*}[t]
    \small
    \renewcommand\arraystretch{1.0}
    \setlength{\abovecaptionskip}{0pt}
	\setlength{\tabcolsep}{0.015\linewidth}
	\newcommand{\classfreq}[1]{{~\tiny(\nuscenesfreq{#1}\%)}}
    \caption{\textbf{Quantitative results on Occ3D-nuScenes.}
    We report mIoU, mIoU$_D$ (dynamic objects only), inference latency (measured on a NVIDIA A100 GPU), and memory consumption.
    Methods with near real-time latency ($\approx$100 ms) are highlighted in gray.
    ($\dagger$ denotes reproduced without CBGS~\cite{cbgs} under the same setting as ours.)}
    \vspace{-2.7em}
    \begin{center}
        \resizebox{\linewidth}{!}{
	\begin{tabular}{l|c|c|c|cc|cc}
		\toprule
        Method & Backbone & Image Size & Visible Mask  & mIoU $\uparrow$ & mIoU$_{D}$ $\uparrow$ & Latency (ms) & Memory (MB) \\
		\midrule
        BEVFormer~\cite{bevformer}& ResNet-101& $900\times1600$ &  \ding{52} & 39.3 & 37.2 & 212.7 & 6,651 \\
        BEVDet4D~\cite{bevdet4d} & ResNet-50 & $256\times704$  & \ding{52}  & 39.2 & 32.8 & 1250.0 & 6,053 \\
        PanoOcc~\cite{panoocc} & ResNet-101&$864\times1600$ & \ding{52}  & 41.6 & 37.3 & 333.3 & 11,991\\
        GEOcc~\cite{geocc} & ResNet-50 & $256\times704$  & \ding{52}  & 43.6 & 38.6 & 450.0 & - \\
        COTR~\cite{cotr}& ResNet-50&$256\times704$   & \ding{52}  & 44.5 & 39.5 & 1111.1 & 10,453\\
        ALOcc$^{\dagger}$~\cite{alocc}& ResNet-50&$256\times704$   & \ding{52}  & 45.0 & 38.7 & 166.7 & 10,793\\
        STCOcc~\cite{stcocc}& ResNet-50&$256\times704$   & \ding{52}  & 44.6 & 40.0 & 212.8 & 7,739\\
        \midrule\midrule
        \rowcolor[gray]{0.95} SparseOcc ~\cite{sparseocc} & ResNet-50 & $256\times704$  & \ding{56} & 30.9 & 28.2 & 56.5  & 6,883 \\
        \rowcolor[gray]{0.95} OPUS ~\cite{opus} & ResNet-50 & $256\times704$ & \ding{56} & 35.6 & 32.6 & 75.7  & 6,735 \\
        \rowcolor[gray]{0.95} FastOcc~\cite{fastocc}& ResNet-101&$320\times800$ & \ding{52}  & 37.2 & - & 93.4 &  -\\        
        \rowcolor[gray]{0.95} ProtoOcc~\cite{protoocc_cvpr}  & ResNet-50 & $432\times800$ & \ding{52} & 37.8 & 32.1 & 105.0 & - \\     
        \rowcolor[gray]{0.95} FB-OCC~\cite{fbocc} &ResNet-50 & $256\times704$   & \ding{52}  & 39.1 & 34.3 & 97.1 & 9,632 \\        
        \rowcolor[gray]{0.95} ProtoOcc~\cite{protoocc_aaai} & ResNet-50 & $256\times704$ & \ding{52} & 39.6 & 34.8 & 77.9 & - \\    
        \rowcolor[gray]{0.95} GSD-Occ ~\cite{gsdocc}& ResNet-50 & $256\times704$ & \ding{52} & 39.4 &  35.1   & 50.0 & 4,759 \\
        \rowcolor[gray]{0.95} ViewFormer$^{\dagger}$ ~\cite{viewformer} & ResNet-50 & $256\times704$  & \ding{52} & 39.6 & 33.3 & 102.0  & 3,103 \\        
        \rowcolor[gray]{0.95} ALOcc-mini$^{\dagger}$ ~\cite{alocc}& ResNet-50 & $256\times704$ & \ding{52} & 40.6 &  35.6  & 33.1 & 2,577 \\
        \midrule
        \rowcolor[gray]{0.95} StreamOcc (Ours) & ResNet-50 & $256\times704$  & \ding{52}  & \textbf{41.9} & \textbf{38.1} & 83.3 & 2,788 \\
		\bottomrule
	\end{tabular}
     }
\vspace{-2.em}    
\label{tab:nuscenes_sota}
\end{center}
\end{table*}

\begin{table}[t]
\centering
\setlength{\tabcolsep}{0.45mm}

\begin{minipage}[t][3cm][t]{0.53\linewidth}
\centering
\renewcommand{\arraystretch}{1.0}
\scriptsize
\setlength{\tabcolsep}{0.46pt}
\caption{\textbf{Results on the SurroundOcc dataset~\cite{surroundocc}.} We report IoU, mIoU, mIoU$_D$, inference latency (Lat., ms on a NVIDIA 4090 GPU) and memory consumption (Mem., MB).}
\label{tab:surround_sota}
\vspace{-1.em}
\begin{tabular}{l|c|c|c|@{\hspace{2pt}}c@{\hspace{2pt}}|@{\hspace{1.5pt}}c}
\toprule
Method\rule{0pt}{2.8ex} & {\tiny IoU}$\uparrow$ & {\tiny mIoU}$\uparrow$ & {\tiny mIoU$_{D}$}$\uparrow$ & {\tiny Lat.} & {\tiny Mem.}  \\
\midrule
TPVFormer~\cite{tpvformer} & 30.9 & 17.1 & 13.4 & 320 & 5,100  \\
SurroundOcc~\cite{surroundocc} & 31.5 & 20.3 & 16.0 & 344 & 5,491  \\
GaussianFormer~\cite{gaussianformer_v1} & 29.8 & 19.1 & 16.7 & 372 & 6,229 \\
GaussianFormer-2~\cite{gaussianformer_v2} & 31.7 & 20.8 & 18.3 & 451 & 4,535  \\
QuadricFormer~\cite{quadricformer} & 32.1 & 21.1 & 17.3 & 179 & 2,563  \\
GaussianWorld~\cite{gaussianworld} & 33.0 & 21.9 & 19.0 & 228 & 7,030 \\\midrule
\rowcolor[gray]{0.95} StreamOcc (Ours) & \textbf{33.8} & \textbf{23.4} & \textbf{21.0} & \textbf{84} & 2,788  \\
\bottomrule
\end{tabular}
\end{minipage}
\hfill
\begin{minipage}[t][6.2cm][t]{0.44\linewidth}
\centering
\renewcommand{\arraystretch}{1.02}
\scriptsize
\caption{\textbf{RayIoU~\cite{sparseocc} evaluation on Occ3D-nuScenes~\cite{occ3d}.} Comparison with real-time methods ($\leq$100 ms), without visibility mask.}
\label{tab:rayiou}
\vspace{-1.em}
\setlength{\tabcolsep}{0.3mm}
\begin{tabular}{l|c|c|c|c}
\toprule
Method\rule{0pt}{2.4ex} & RayIoU & \textit{1m} & \textit{2m} & \textit{4m} \\
\midrule
SpaseOcc~\cite{sparseocc} & 35.1 & 29.1 & 35.8 & 40.3 \\
FB-Occ~\cite{fbocc} & 35.6 & - & - & - \\
GSD-Occ~\cite{gsdocc} & 38.9 & 33.0 & 39.7 & 44.1 \\
ODG~\cite{odg} & 39.2 & - & - & - \\
ALOcc-mini~\cite{alocc} & 39.3 & 32.9 & 40.1 & 44.8 \\
OPUS~\cite{opus} & 40.3 & 33.7 & 41.1 & 46.0 \\\midrule
StreamOcc (Ours) & \textbf{41.1} & \textbf{34.2} & \textbf{41.9} & \textbf{47.1} \\
\bottomrule
\end{tabular}
\end{minipage}
\vspace{-6.4em}
\end{table}

\subsection{Comparison with SOTA Methods}
To comprehensively compare StreamOcc with prior works, we evaluate it on Occ3D-nuScenes~\cite{occ3d} against multi-frame fusion approaches, additionally reporting RayIoU~\cite{sparseocc} for ray-wise consistency, and on the SurroundOcc benchmark~\cite{surroundocc} against Gaussian- and Quadric-based methods. Furthermore, since accurate modeling of dynamic objects is safety-critical in autonomous driving, we report mIoU$_D$ to explicitly quantify performance on dynamic objects.

\myparagraph{Comparison with SOTA Methods on Occ3D-nuScenes.}
We first evaluate StreamOcc on the Occ3D-nuScenes dataset~\cite{occ3d} in Tab.~\ref{tab:nuscenes_sota}. 
StreamOcc achieves 41.9 mIoU and 38.1 mIoU$_D$, the best accuracy among real-time methods. 
It outperforms the prior-best real-time method, ALOcc-mini~\cite{alocc}, by +1.3 mIoU and +2.5 mIoU$_D$. 
StreamOcc also surpasses multi-frame fusion methods that use sparse or compressed representations, such as SparseOcc~\cite{sparseocc}, FastOcc~\cite{fastocc}, OPUS~\cite{opus}, and GSD-Occ~\cite{gsdocc}. 
Compared to ViewFormer~\cite{viewformer}, which uses compressed multi-frame features propagated in a streaming manner, StreamOcc achieves higher accuracy (+2.3 mIoU and +4.8 mIoU$_D$) while operating faster. 

Compared to multi-frame dense voxel fusion methods, StreamOcc achieves stronger performance with significantly lower cost than PanoOcc~\cite{panoocc}, which is about $4\times$ slower and uses $4.3\times$ more memory. Although COTR~\cite{cotr} and the ALOcc~\cite{alocc} attain higher accuracy, they incur substantially higher overhead, being about $13\times$ and $2\times$ slower than StreamOcc while consuming $3.7\times$ and $3.8\times$ more memory, respectively. Such high latency and memory overhead make these approaches less suitable for on-vehicle deployment.

\myparagraph{Comparison with SOTA Methods on SurrondOcc-benchmark.}
On the SurroundOcc~\cite{surroundocc} dataset (Tab.~\ref{tab:surround_sota}), StreamOcc likewise achieves the best performance across IoU, mIoU, and mIoU$_D$, outperforming Gaussian- and Quadric-based approaches~\cite{gaussianformer_v1,gaussianformer_v2,quadricformer}. 
Notably, it also surpasses the prior best method, GaussianWorld~\cite{gaussianworld}, by +1.5 mIoU and +2.0 mIoU$_D$, while requiring substantially lower cost, running 2.7$\times$ faster and using 2.5$\times$ less memory.
 
\myparagraph{Comparison with SOTA Methods on RayIoU.} We further evaluate methods using RayIoU~\cite{sparseocc} on Occ3D-nuScenes~\cite{occ3d} (Tab.~\ref{tab:rayiou}), which measures the consistency of occupancy prediction along viewing rays. StreamOcc achieves the highest overall RayIoU (41.1), consistently outperforming prior real-time approaches. Together, these results demonstrate that StreamOcc provides state-of-the-art accuracy while maintaining practical runtime and memory efficiency, making it suitable for real-world deployment.

\begin{table*}[t]
\centering
\footnotesize
\setlength{\tabcolsep}{1mm}
\caption{\textbf{Ablation study on the Occ3D-nuScenes dataset.} Evaluates the effect of each component of StreamOcc in terms of overall mIoU (mIoU$_{A}$), dynamic and static object mIoU (mIoU$_{D}$ and mIoU$_{S}$, respectively), and per-class performance.}
\vspace{-1.em}
\label{tab:ablation}
\resizebox{\textwidth}{!}{
\begin{tabular}{l|c|c|c|cccccccccc|cccccc}
\toprule
&&&
& \multicolumn{10}{c|}{mIoU $\uparrow$ (Dynamic Objects) }
& \multicolumn{6}{c}{mIoU $\uparrow$ (Static Objects)} \\
\cmidrule{5-20}
Method
& \begin{sideways}{$\text{mIoU}_{A} \uparrow$}\end{sideways}
& \begin{sideways}{$\text{mIoU}_{D} \uparrow$}\end{sideways}
& \begin{sideways}{$\text{mIoU}_{S} \uparrow$}\end{sideways}
& \begin{sideways}{\textcolor{barriercolor}{$\blacksquare$} \texttt{barrier}}\end{sideways}  
& \begin{sideways}{\textcolor{bicyclecolor}{$\blacksquare$} \texttt{bicycle}}\end{sideways}  
& \begin{sideways}{\textcolor{buscolor}{$\blacksquare$} \texttt{bus}}\end{sideways}  
& \begin{sideways}{\textcolor{carcolor}{$\blacksquare$} \texttt{car}}\end{sideways}  
& \begin{sideways}{\textcolor{constructcolor}{$\blacksquare$} \texttt{cons. veh.}}\end{sideways}  
& \begin{sideways}{\textcolor{motorcolor}{$\blacksquare$} \texttt{motorcycle}}\end{sideways}  
& \begin{sideways}{\textcolor{pedestriancolor}{$\blacksquare$} \texttt{pedestrian}}\end{sideways}  
& \begin{sideways}{\textcolor{trafficcolor}{$\blacksquare$} \texttt{traffic cone}}\end{sideways}  
& \begin{sideways}{\textcolor{trailercolor}{$\blacksquare$} \texttt{trailer}}\end{sideways}  
& \begin{sideways}{\textcolor{truckcolor}{$\blacksquare$} \texttt{truck}}\end{sideways}  
& \begin{sideways}{\textcolor{drivablecolor}{$\blacksquare$} \texttt{drive. surf.}}\end{sideways}  
& \begin{sideways}{\textcolor{otherflatcolor}{$\blacksquare$} \texttt{other flat}}\end{sideways}  
& \begin{sideways}{\textcolor{sidewalkcolor}{$\blacksquare$} \texttt{sidewalk}}\end{sideways}  
& \begin{sideways}{\textcolor{terraincolor}{$\blacksquare$} \texttt{terrain}}\end{sideways}  
& \begin{sideways}{\textcolor{manmadecolor}{$\blacksquare$} \texttt{manmade}}\end{sideways}  
& \begin{sideways}{\textcolor{vegetationcolor}{$\blacksquare$} \texttt{vegetation}}\end{sideways}
\\
\midrule
A. Base (Single-Frame) & 36.8 & 32.6 & 48.5 & 44.1 & 22.4 & 44.9 & 50.0 & 21.2 & 23.9 & 25.7 & 25.4 & 32.0 & 36.3 & 78.9 & 41.0 & 48.5 & 51.4 & 38.1 & 32.9  \\ 
B. StreamAgg & 40.4 & 35.4 & \textbf{53.4} & 47.2 & 26.2 & 45.5 & 53.8 & 24.8 & 29.4 & 26.9 & 29.5 & 32.8 & 38.4 & 83.2 & 45.6 & \textbf{54.0} & \textbf{57.6} & \textbf{42.7} & \textbf{37.5}  \\ 
C. B + Detection Head & 40.8 & 36.3 & 53.1  & 48.0 & 26.8 & 46.0 & 54.6 & 24.9 & 29.6 & 28.6 & 30.7 & 33.2 & 40.9 & 83.2 & 45.4 & 53.8 & 57.3 & 41.8 & 37.4 \\ 
D. B + QueryAgg (Ours) & \textbf{41.9} & \textbf{38.1} & 53.3  & \textbf{50.8} & \textbf{29.6} & \textbf{48.5} & \textbf{56.1} & \textbf{25.4} & \textbf{31.4} & \textbf{30.0} & \textbf{33.0} & \textbf{34.6} & \textbf{41.9} & \textbf{83.3} & \textbf{45.9} & 53.8 & 57.3 & 42.3 & 37.2 \\  \midrule
\end{tabular}
}
\vspace{-2.0em}
\end{table*}

\subsection{Ablations}
In this section, we conduct ablation studies on Occ3D-nuScenes to analyze the contribution of each module and validate our design choices. Inference latency and GPU memory are measured on one NVIDIA A100.

\myparagraph{Effect of StreamOcc Components.}
Tab.~\ref{tab:ablation} presents an ablation study on the Occ3D-nuScenes dataset~\cite{occ3d}. Base model (A), using only a single-frame, achieves 36.8 mIoU. Adding StreamAgg (B), which effectively rectifies warping distortions to maintain temporal consistency while sequentially accumulating voxel features over time, we observe a substantial performance improvement to 40.4 mIoU. However, (B) still shows a large performance gap between static and dynamic objects (53.4 vs. 35.4), indicating that voxel-only accumulation struggles to capture the representations of dynamic objects.
To address this limitation, we add an auxiliary detection head in (C) for dynamic-object supervision. This yields only a marginal gain (+0.9 mIoU on dynamic objects), proving that indirect supervision lacks the capacity to enrich these representations. In contrast, our QueryAgg module (D) directly injects instance-level dynamic object features into the corresponding voxel regions, yielding a substantial gain of +2.7 mIoU$_{D}$ over (B). Overall, the combination of StreamAgg and QueryAgg achieves the best performance (41.9 mIoU), strengthening dynamic object representations while maintaining prediction accuracy for static objects.

\myparagraph{Ablation of Refinement Module.}
Tab.~\ref{tab:streamagg_ablation} analyzes the effect of our Adaptive Residual Refinement module and its supervision. 
Naive voxel streaming yields sub-optimal result because the propagated voxel features suffer from warping-induced distortions during temporal alignment. 
Adopting Adaptive Residual Refinement (w/o supervision) alleviates these artifacts, improving performance to 39.84 mIoU (+1.12 mIoU) with minimal overhead (only +5 ms latency and +14 MB memory). 
Semantic Supervision further enhances temporal semantic consistency, while Geometry Supervision guides the refinement to focus on informative features, achieving the best performance of 40.37 mIoU without additional computational cost. 
Overall, each component contributes complementary gains, and the full StreamAgg provides a stable and spatially coherent voxel representation.

\begin{table}[t]
\small
\begin{minipage}{0.55\columnwidth}
\scriptsize
\vspace{0pt}
\centering
\renewcommand{\arraystretch}{1.1}
\setlength{\tabcolsep}{0.5mm}
\caption{\textbf{Refinement ablation.}}\label{tab:streamagg_ablation}\vspace{-1em}
\begin{tabular}{l|ccc}
    \toprule
    Method & mIoU$\uparrow$ & Lat. & Mem. \\
    \midrule
    Naive Voxel Streaming & 38.72 & 43.9 & 2,423 \\ 
    (+) Adaptive Residual Refinement & 39.84 & 49.0 & 2,437  \\ 
    (+) Semantic Supervision & 40.25 & 49.0 & 2,437 \\
    (+) Geometry Supervision & \textbf{40.37} & 49.0 & 2,437 \\
    \bottomrule
\end{tabular}
\end{minipage}
\begin{minipage}{0.44\columnwidth}
\centering
\scriptsize
\renewcommand{\arraystretch}{0.93}
\setlength{\tabcolsep}{0.7mm}
\caption{\textbf{QueryAgg ablation.}}\label{tab:detection_ablation}\vspace{-1em}
\begin{tabular}{ccc|ccc}
\toprule
 I2Q & V2Q & DQA & mIoU$\uparrow$ & NDS$\uparrow$ & mAP$\uparrow$ \\
\midrule
  &  &  & 40.37 & -- & -- \\
  & \CheckmarkBold &  & 40.06 & 0.4704 & 0.3458 \\
\CheckmarkBold &  &  & 40.42 & 0.4904 & 0.3516 \\
 \CheckmarkBold & \CheckmarkBold &  & 40.78 & 0.4964 & 0.3620 \\
 \CheckmarkBold & \CheckmarkBold & \CheckmarkBold & \textbf{41.90} & \textbf{0.5001} & \textbf{0.3681} \\
\bottomrule
\end{tabular}
\end{minipage}
\vspace{.8em}
\caption{\textbf{Image-to-voxel aggregation ablation.}}
\label{tab:image_to_voxel}
\vspace{-.7em}
\scriptsize
\centering
\begin{tabular}{l|ccc|cc}
    \toprule
    Model & mIoU$_\text{A}$ $\uparrow$ & mIoU$_\text{D}$ $\uparrow$ & mIoU$_\text{S}$ $\uparrow$ & Latency (ms) & Memory (MB) \\
    \midrule
    StreamAgg only & 40.37 & 35.36 & 53.42 & 49.0 & 2,437 \\  
    StreamAgg + Spatial Cross-Attention & 40.72 & 35.77 & \textbf{53.59} & 95.2 & 3,554 \\ 
    StreamAgg + QueryAgg & \textbf{41.90} & \textbf{38.12} & 53.31 & 83.3 & 2,788 \\
    \bottomrule
\end{tabular}
\vspace{-1.5em}
\end{table}

\myparagraph{How to Utilize a Detector for Joint Learning?}
Tab.~\ref{tab:detection_ablation} presents an ablation study on how QueryAgg components improve both detection and occupancy prediction. The StreamAgg-only setting serves as the baseline. 
Applying a query-based detector on voxel features (V2Q) for multi-task learning, as in~\cite{panoocc,occnet}, is inefficient due to the large search space and may fail to detect objects that are weakly represented in voxel features. 
Using an Image-to-Query detector (I2Q) provides finer image-based detection, but yields limited gains without voxel-level guidance. 
Combining I2Q with V2Q improves this by coupling semantic cues from images with geometric details from voxel features to guide dynamic object localization, yet still lacks direct voxel interaction. 
The complete StreamOcc further introduces direct query--voxel interaction by injecting instance-level features into voxel features, complementing dynamic object representations. This interactive design enhances both detection and occupancy prediction, which are built upon distinct representations, achieving the highest NDS, mAP, and mIoU.

\myparagraph{Effect of Query-guided Aggregation.}
Tab.~\ref{tab:image_to_voxel} compares image-to-voxel aggregation methods for overcoming the limitations of voxel-based encoding. Prior works adopt Spatial Cross-Attention~\cite{occformer,cotr,geocc} to distribute image features across all voxel grids, providing slight improvements. However, spatial misalignment between voxel and image space often leads to hallucinated mappings~\cite{voxformer}, and the lack of explicit supervision on what to extract and where to place it results in suboptimal performance. In contrast, Query-guided Aggregation (QueryAgg) explicitly extracts dynamic object features and selectively integrates them into the voxel regions they occupy. This targeted aggregation strategy mitigates hallucinations, reduces unnecessary computation, and yields a +2.76 mIoU gain on dynamic objects and +1.53 overall, while maintaining high efficiency with 83.3 ms latency and 2,788 MB memory.

\begin{figure*}[t]
  \centering
    \includegraphics[width=.98\linewidth]{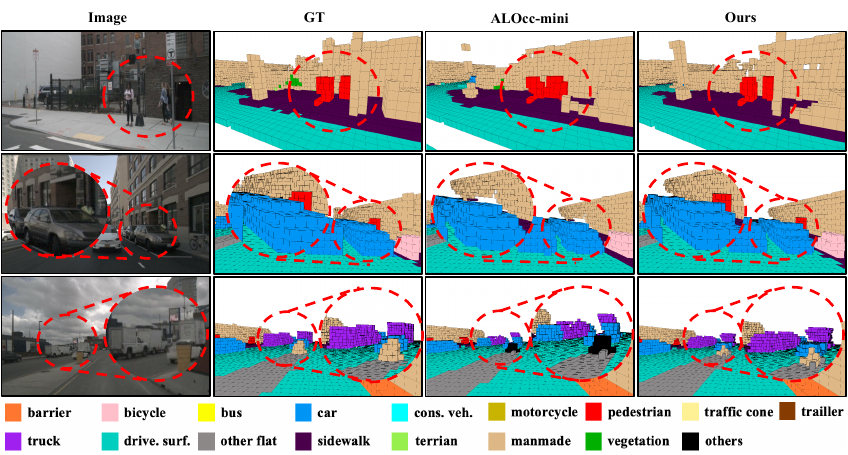}
  \vspace{-.5em}
  \caption{\textbf{Qualitative comparison with other SOTA method.} StreamOcc enables more precise reconstruction of dynamic objects and more accurate semantic predictions for static objects than previous best real-time method, ALOcc-mini~\cite{alocc}.}
  \label{fig:visualization_main}
  \vspace{-2.0em}
\end{figure*}

\begin{figure}[t]
  \centering  \includegraphics[width=.98\linewidth]{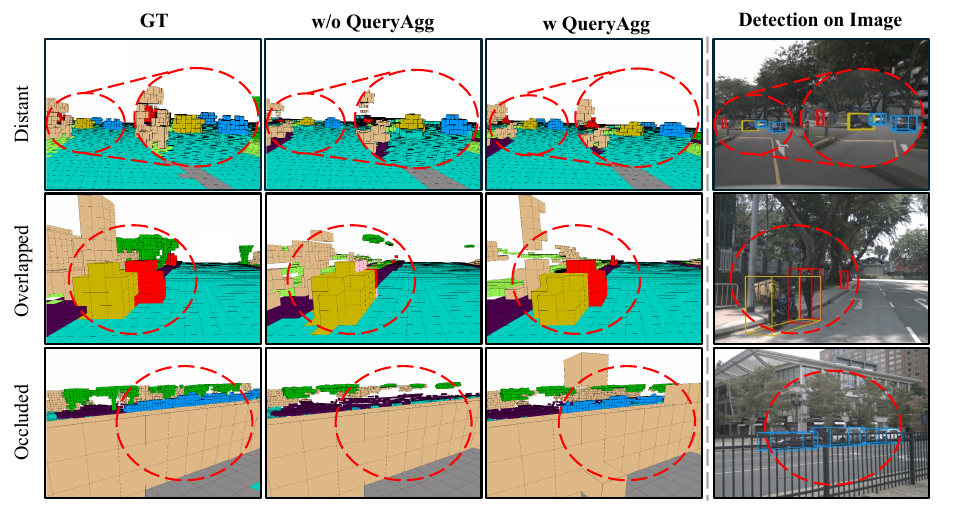}
  \vspace{-1.2em}
  \caption{\textbf{Effect of QueryAgg}. QueryAgg enables more accurate predictions for distant, overlapped, and occluded dynamic objects. The fourth column shows instance-query detections output used in QueryAgg, and red dotted circles highlight improved regions.}
  \label{fig:visualization}
  \vspace{-2.em}
\end{figure}
\subsection{Qualitative Analysis}
In Fig.~\ref{fig:visualization_main}, we compare StreamOcc with the previous state-of-the-art real-time method, ALOcc-mini~\cite{alocc}. 
StreamOcc produces predictions closer to the ground truth in the regions highlighted by red dotted circles. 
It reconstructs pedestrians more precisely, whereas ALOcc-mini produces overly smoothed pedestrian groups (Row~1), correctly captures even a partially occluded pedestrian that is missed by ALOcc-mini (Row 2), and better reconstructs distant moving vehicles with shapes closer to the ground truth, while also improving semantic predictions not only for dynamic objects but also for static objects (Row 3). 
Overall, these qualitative results support our quantitative findings that StreamOcc strengthens dynamic-object reconstruction while maintaining accurate semantics for static objects under real-time constraints.

Additionally, Fig.~\ref{fig:visualization} shows that QueryAgg effectively mitigates information loss induced by image-to-voxel projection, enhancing voxel representations of dynamic objects that appear small due to long-range distance, overlap with nearby instances, or are partially occluded. Compared to the model without QueryAgg, our method with QueryAgg yields predictions closer to the ground truth in these challenging scenarios. As illustrated in the rightmost column, the instance queries accurately detect dynamic objects, and QueryAgg performs targeted aggregation by injecting instance-level features from these dynamic queries into the voxel regions occupied by the corresponding objects. This design enables voxel features to retain a more precise representation of dynamic objects even under sparse projections for distant objects, severe overlap that causes feature mixing in coarse grids, and occlusions that lead to incomplete projections.

\section{Conclusion}
\label{sec:conclusion}
In this paper, we presented StreamOcc, a novel framework that utilizes dense voxel streaming for efficient and accurate 3D occupancy prediction. 
StreamOcc addresses the challenges of naive dense voxel streaming by introducing two aggregation strategies: StreamAgg, which enables temporally consistent streaming by mitigating warping-induced distortions in propagated voxel features, and QueryAgg, which enhances dynamic object representations by selectively injecting instance-level semantics into corresponding occupied voxel regions. 
Extensive experiments on Occ3D-nuScenes and SurroundOcc-benchmark demonstrate that StreamOcc achieves state-of-the-art performance, outperforming prior methods that utilize sparse representation while maintaining computational costs. 
Overall, StreamOcc unlocks dense voxel streaming as a practical paradigm and establishes a strong framework for future research in dense 3D scene understanding.

\clearpage
\appendix
\setcounter{figure}{0}
\setcounter{table}{0}
\renewcommand{\thefigure}{S\arabic{figure}}
\renewcommand{\thetable}{S\arabic{table}}
\renewcommand{\theHfigure}{supp.figure.\arabic{figure}}
\renewcommand{\theHtable}{supp.table.\arabic{table}}
\renewcommand{\theHsection}{appendix.\Alph{section}}
\renewcommand{\theHsubsection}{appendix.\Alph{section}.\arabic{subsection}}
\renewcommand{\theHsubsubsection}{appendix.\Alph{section}.\arabic{subsection}.\arabic{subsubsection}}
\begin{center}
{\Large\bfseries Supplementary Material}\par
\end{center}
\section{Further Implementation Details}
\label{sec:further_implement}
\subsection{Lightweight 3D-FPN}

Following BEVDepth~\cite{bevdepth}, we lift image features into the 3D voxel space, obtaining $\mathbf{V}_{\text{init}} \in \mathbb{R}^{C \times X \times Y \times Z}$, which encodes both spatial and semantic information and serves as the foundation for 3D occupancy prediction. To improve computational efficiency while preserving essential structure, we apply a lightweight 3D-FPN that downsamples voxel features, avoiding excessively fine voxel grids that introduce unnecessary computation.

Specifically, $\mathbf{V}_{\text{init}}$ is passed through a Conv3D block to generate multi-scale voxel features: $\mathbf{V}_{1} \in \mathbb{R}^{C_1 \times \frac{X}{2} \times \frac{Y}{2} \times \frac{Z}{2}}$ and $\mathbf{V}_{2} \in \mathbb{R}^{C_2 \times \frac{X}{4} \times \frac{Y}{4} \times \frac{Z}{4}}$. These features, together with $\mathbf{V}_{\text{init}}$, are then uniformly resampled to a resolution of $\frac{X}{2} \times \frac{Y}{2} \times \frac{Z}{2}$ via trilinear interpolation. The interpolated volumes are concatenated and compressed using a Conv1D layer to obtain a compact representation $\mathbf{V}_{\text{down}} \in \mathbb{R}^{C \times \frac{X}{2} \times \frac{Y}{2} \times \frac{Z}{2}}$.
This design substantially reduces computational overhead while maintaining spatial and semantic detail.

\subsection{Query Selection for Dynamic Query Aggregation}
Selecting reliable instance queries is critical for Dynamic Query Aggregation (DQA). 
Naively selecting queries solely based on IoU with ground-truth boxes may introduce shortcut learning, as queries that are spatially close to ground-truth can be mapped into voxel features even when their detection quality is poor. This may implicitly expose ground-truth information during training and degrade generalization at inference.
To address this issue, we adopt a training-time query filtering strategy that combines confidence scores with IoU criteria, and geometric constraints. For sufficiently large objects, we apply an IoU-based rule and select queries that satisfy:
\[
\text{IoU}(\hat{b}, b) > 0.4 \quad \text{and} \quad s > 0.3,
\]
where $\hat{b}$ and $b$ denote the predicted and ground-truth bounding boxes and $s$ is the query confidence score.

For small objects, IoU becomes unstable because even minor deviations in center position or box size can significantly reduce IoU despite accurate detection. Therefore, we introduce a geometry-based criterion that measures the deviation between predicted and ground-truth box geometry. A query is selected if it satisfies:
\[
(\sigma_{c} D_{\text{center}} + \sigma_{b} D_{\text{size}}) < 1.5
\quad \text{and} \quad
s > 0.3,
\]
where $D_{\text{center}}$ denotes the center distance between $\hat{b}$ and $b$, and $D_{\text{size}}$ measures the difference in box dimensions. The weighting coefficients are empirically set to $\sigma_{c}=2.0$ and $\sigma_{b}=1.0$. This criterion allows queries that capture object geometry accurately to be selected even when IoU is unreliable.

During inference, the filtering strategy is simplified by selecting only instance queries with confidence score $s>0.3$, consistent with the detection threshold used in the detector.

\subsection{Auxiliary Supervision Head for Refinement}
To ensure that the Adaptive Residual Refinement module reliably learns to adjust voxel features, we incorporate two auxiliary supervision heads used only during training.

First, for \textbf{Geometry Supervision}, we upsample the spatial attention feature $\mathbf{M}_s$ to the original voxel resolution and apply the MLP-based decoder to predict a binary occupancy map $\mathbf{V}^t_\text{geo} \in \mathbb{R}^{1 \times X \times Y \times Z}$. 
This prediction is supervised with the ground-truth occupied mask via binary cross-entropy loss, encouraging the refinement module to focus on informative features that are meaningful for reducing distortions.
Additionally, we apply \textbf{Semantic Supervision} to the refined warped voxel feature $\mathbf{V}_{\text{refwarp}}^t$. 
Specifically, $\mathbf{V}_{\text{refwarp}}^t$ is decoded using the same upsampling and MLP-based decoder to predict the voxel semantic occupancy at the current timestep, $\mathbf{V}^t_\text{sem} \in \mathbb{R}^{Class \times X \times Y \times Z}$. 
This supervision explicitly guides the refinement module to correct warping-induced distortions and align the refined representation with the current scene state. 

Since this refinement is applied to the propagated voxel features before they are combined with the current voxel features, it helps maintain temporal consistency and enables stable temporal accumulation across timesteps.

\subsection{Loss Functions}
To train our model, we combine multiple task-specific loss functions. 
For depth estimation, we adopt the depth loss $\mathcal{L}_{\text{depth}}$ from BEVDepth~\cite{bevdepth}. 
Voxel occupancy prediction is supervised using the cross-entropy-based occupancy loss $\mathcal{L}_{\text{occ}}$, applied to the predictions generated from the final voxel features $V_{\text{fin}}$ via an MLP decoder. 
For the Image-to-Query Detector, we employ the detection loss $\mathcal{L}_{\text{det}}$ from Sparse4D v3~\cite{sparse4dv3}. 
To further enhance voxel representations, we incorporate an Auxiliary Mask Decoder with the mask loss $\mathcal{L}_{\text{mask}}$.

In addition, the Semantic Head supervises the refined warped voxel features by predicting the current semantic occupancy distribution, which is optimized using a cross-entropy loss $\mathcal{L}_{\text{sem}}$. 
The Geometry Head predicts whether each voxel is occupied or empty and is supervised with a binary cross-entropy loss $\mathcal{L}_{\text{geo}}$.

The overall training objective $\mathcal{L}_{\text{total}}$ is defined as:
\vspace{-.3em}
\begin{equation}
\begin{aligned}
\mathcal{L}_{\text{total}} ={} &
\lambda_{\text{depth}} \cdot \mathcal{L}_{\text{depth}}
+ \lambda_{\text{occ}} \cdot \mathcal{L}_{\text{occ}}
+ \lambda_{\text{det}} \cdot \mathcal{L}_{\text{det}} \\
& + \lambda_{\text{mask}} \cdot \mathcal{L}_{\text{mask}} 
+ \lambda_{\text{sem}} \cdot \mathcal{L}_{\text{sem}}
+ \lambda_{\text{geo}} \cdot \mathcal{L}_{\text{geo}},
\end{aligned}
\end{equation}
where $\lambda_{\text{depth}}$, $\lambda_{\text{occ}}$, $\lambda_{\text{det}}$, $\lambda_{\text{mask}}$, $\lambda_{\text{sem}}$, and $\lambda_{\text{geo}}$ denote the balancing weights for each loss term. 
In our experiments, these weights are empirically set to $0.05$, $10.0$, $0.2$, $1.0$, $10.0$, and $10.0$, respectively.

\section{More Experiments}
\subsection{Evaluation Metrics}
We evaluate occupancy prediction using IoU and mIoU, and further adopt RayIoU as the primary evaluation metric following SparseOcc~\cite{sparseocc}. Unlike voxel-level mIoU, RayIoU measures occupancy prediction along rays, jointly considering semantic correctness and depth accuracy. Specifically, a predicted ray is regarded as a true positive only when its semantic class matches the ground truth and the depth error falls within a predefined threshold (e.g., 1\,m, 2\,m, or 4\,m). 
The metrics are defined as follows:
\begin{equation}
\text{mIoU/RayIoU} = \frac{1}{|C|}\sum_{i \in C}\frac{TP_i}{TP_i + FP_i + FN_i},
\end{equation}
\begin{equation}
\text{IoU} = \frac{TP_{c_0}}{TP_{c_0} + FP_{c_0} + FN_{c_0}},
\end{equation}
where $TP_i$, $FP_i$, and $FN_i$ denote the numbers of true positives, false positives, and false negatives for class $i$, respectively, $C$ is the set of semantic classes, and $c_0$ denotes the occupied class.

\subsection{Ablation on Query Selection Strategy}
\label{sec:query_selection_ablation}

To analyze the impact of the query selection strategy used in Dynamic Query Aggregation (DQA), we conduct an ablation study by varying the criteria used to select instance queries during training. In particular, we compare four strategies: selecting queries based only on IoU with ground-truth boxes, selecting based only on classification confidence, combining IoU and classification scores, and our full strategy that additionally incorporates geometry-based constraints for small objects.

As shown in Table~\ref{tab:query_selection_ablation}, selecting queries based solely on IoU yields limited performance ($39.9$ mIoU), as it may select queries that are spatially close to ground truth but poorly detected. Using only classification confidence slightly improves performance ($40.1$ mIoU), but still lacks reliable spatial alignment. Combining IoU and classification scores significantly improves performance ($41.3$ mIoU) by filtering unreliable detections.

Our full query selection strategy further incorporates geometry-based constraints for small objects, addressing the instability of IoU for small bounding boxes. This strategy achieves the best performance of $\mathbf{41.9}$ mIoU, demonstrating that geometry-aware query filtering effectively improves the reliability of instance queries used in DQA.

\begin{table}[h]
\centering
\small
\setlength{\tabcolsep}{2.mm}
\caption{\textbf{Ablation study on query selection strategy for Dynamic Query Aggregation.}}
\vspace{-.8em}
\label{tab:query_selection_ablation}
\begin{tabular}{ccc|c}
\toprule
IoU & Cls & Geo & mIoU $\uparrow$ \\
\midrule
\CheckmarkBold &  &  & 39.9 \\
 & \CheckmarkBold &  & 40.7 \\
\CheckmarkBold & \CheckmarkBold &  & 41.3 \\
\CheckmarkBold & \CheckmarkBold & \CheckmarkBold & \textbf{41.9} \\
\bottomrule
\end{tabular}
\end{table}

\subsection{More visualization}
\label{sec:more_vis}
\begin{figure*}[t]
  \centering
    \includegraphics[width=.99\linewidth]{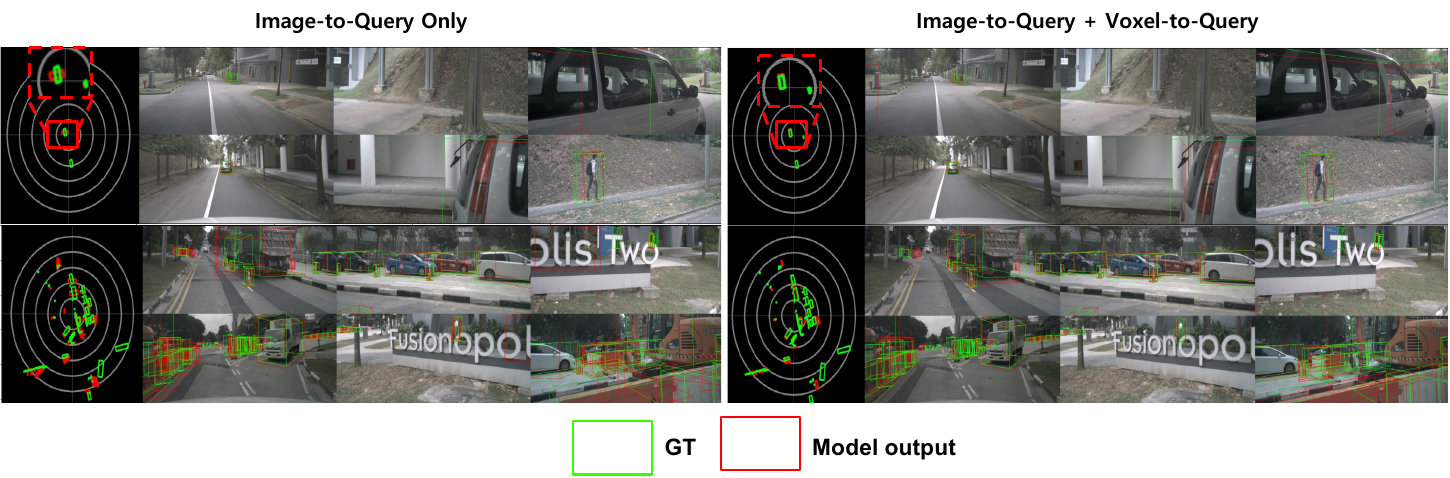}
  \vspace{-1em}
  \caption{\textbf{Qualitative comparison of features used for detection.} The left column shows detection results using only image features, whereas the right column additionally leverages voxel features that provide complementary geometric information. Incorporating voxel-level geometric cues reduces false positives arising along rays and enables more precise 3D detections.}
  \vspace{-1.3em}
  \label{fig:vox2query}
\end{figure*}

\myparagraph{Effect of voxel features on detection.}
Figure~\ref{fig:vox2query} provides a qualitative comparison of the features used for detection. The left column shows detection results using only image features, whereas the right column additionally leverages voxel features that provide complementary geometric information. By incorporating voxel-level geometric cues, the detector reduces false positives arising along rays and produces more precise 3D detections.

\begin{figure*}[h]
  \centering
    \includegraphics[width=.96\linewidth]{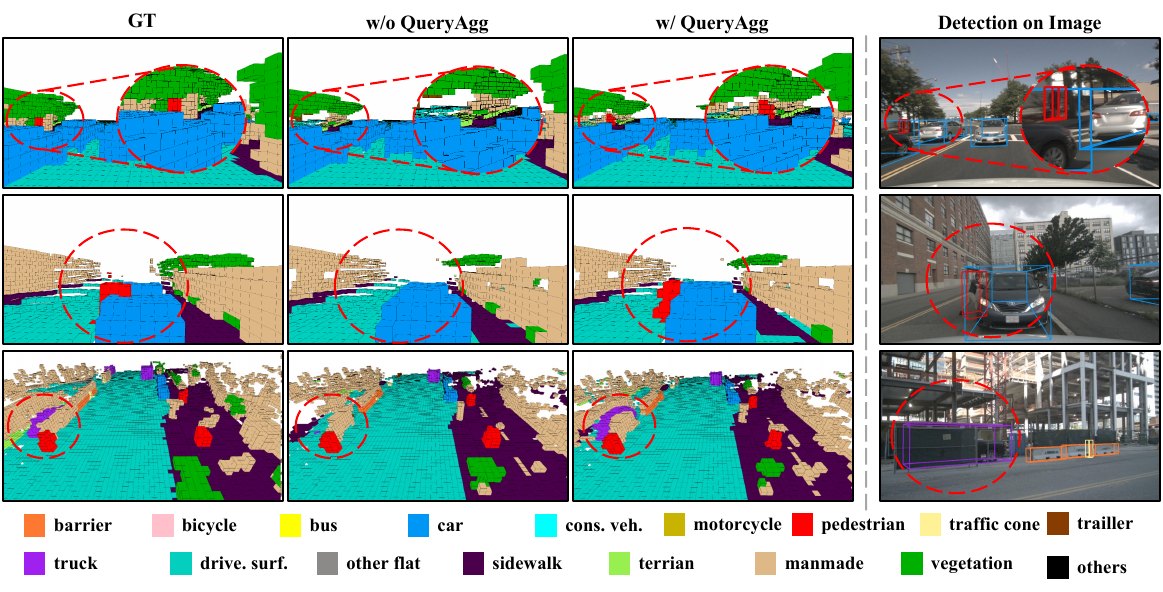}
  \vspace{-1em}
  \caption{\textbf{Visualization results of StreamOcc across three distinct scenarios.} Each row shows the ground-truth (GT), results without QueryAgg (w/o QueryAgg), with QueryAgg (w/ QueryAgg), and detection outputs from the instance queries.}
  \vspace{-1.5em}
  \label{fig:visual_all}
\end{figure*}

\myparagraph{Effect of QueryAgg on occupancy prediction.}
Figure~\ref{fig:visual_all} further demonstrates the effectiveness of StreamOcc across three representative scenarios by comparing the ground truth, results without QueryAgg, results with QueryAgg, and the detection outputs from the instance queries. 
In the first row, QueryAgg enhances the representation of distant pedestrians. 
In the second row, it helps separate nearby objects and more accurately capture their occupancy states. 
In the third row, QueryAgg enables the occupancy map to recover a truck that is partially occluded by a fence. Since the fence obstructs the truck in the image view, its semantic information is not reliably projected into 3D voxel space. By directly aggregating object-level information from instance queries, QueryAgg allows the truck to be more accurately represented in the occupancy map.

\begin{figure*}[t!]
  \centering
    \includegraphics[width=.96\linewidth]{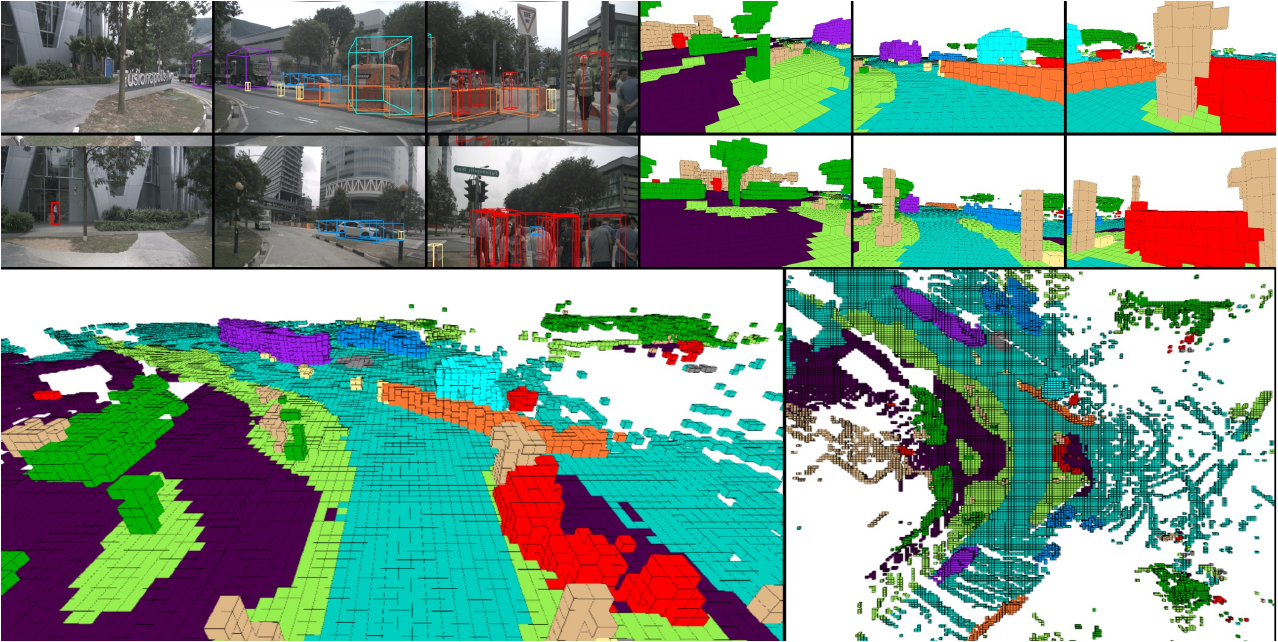}
  \vspace{-1em}
  \caption{\textbf{Visualization of 3D occupancy prediction in a dynamic urban scene with construction and pedestrians.} The top-left shows object detection on the images, the top-right presents occupancy prediction in the camera view, the bottom-left illustrates the top-front view, and the bottom-right depicts the top-down occupancy prediction results.}
  \vspace{-1.5em}
  \label{fig:visual_all_1}
\end{figure*}

\begin{figure*}[t!]
  \centering
    \includegraphics[width=.96\linewidth]{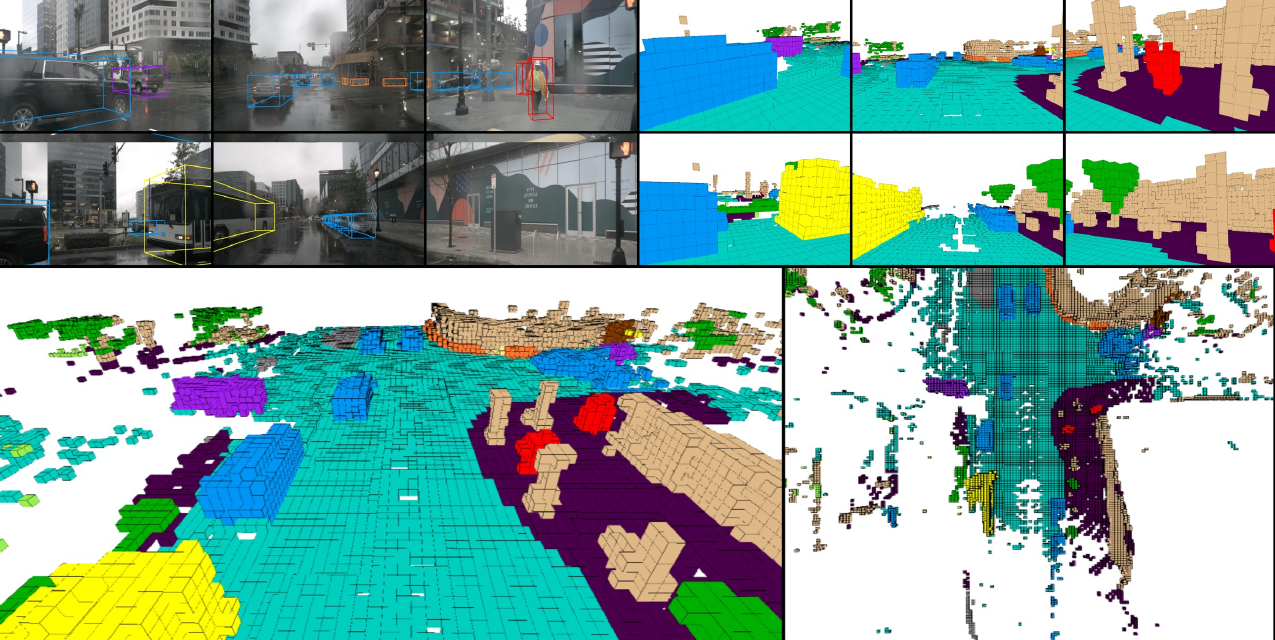}
  \vspace{-1em}
  \caption{\textbf{Visualization of 3D occupancy prediction at a crowded intersection on a rainy day.} The top-left shows object detection in the image space, the top-right presents occupancy prediction in the camera view, the bottom-left illustrates the top-front view, and the bottom-right depicts the top-down occupancy prediction results.}
  \vspace{-4em}
  \label{fig:visual_all_2}
\end{figure*}

\begin{figure*}[t!]
  \centering
    \includegraphics[width=.96\linewidth]{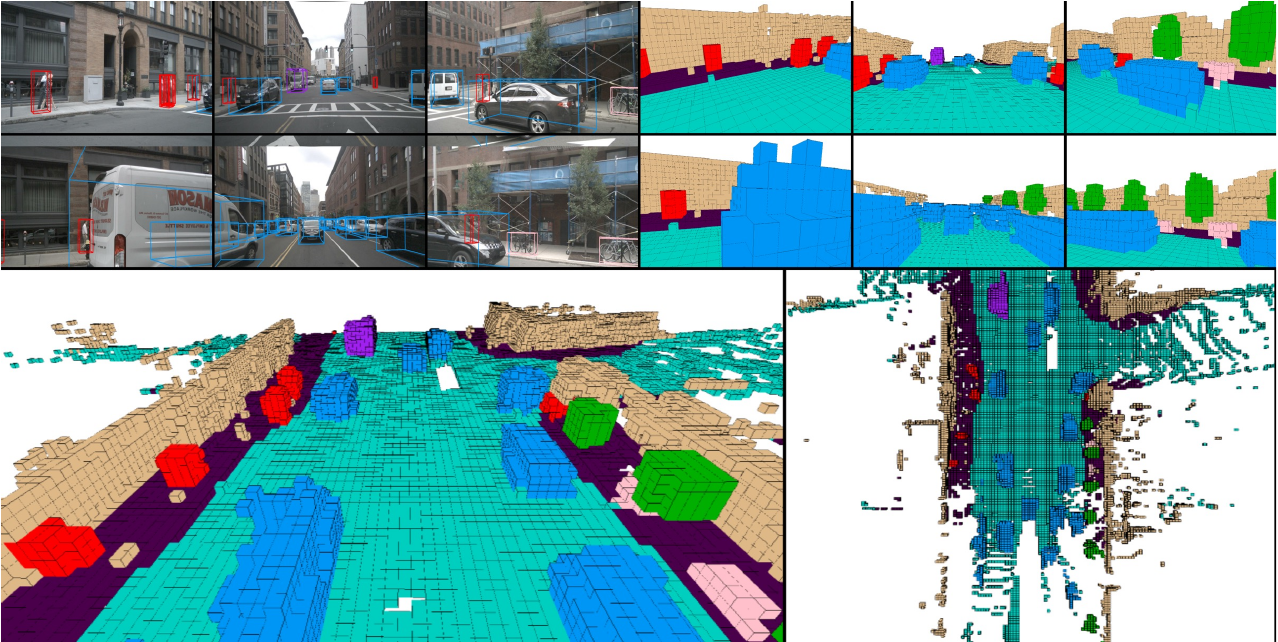}
  \vspace{-1em}
  \caption{\textbf{Visualization of 3D occupancy prediction in a narrow urban street with parked and moving vehicles, pedestrians, and bicycles.} The top-left shows object detection in the image space, the top-right presents occupancy prediction in the camera view, the bottom-left illustrates the top-front view, and the bottom-right depicts the top-down occupancy prediction results.}
  \vspace{-1.5em}
  \label{fig:visual_all_3}
\end{figure*}

\begin{figure*}[t!]
  \centering
    \includegraphics[width=.96\linewidth]{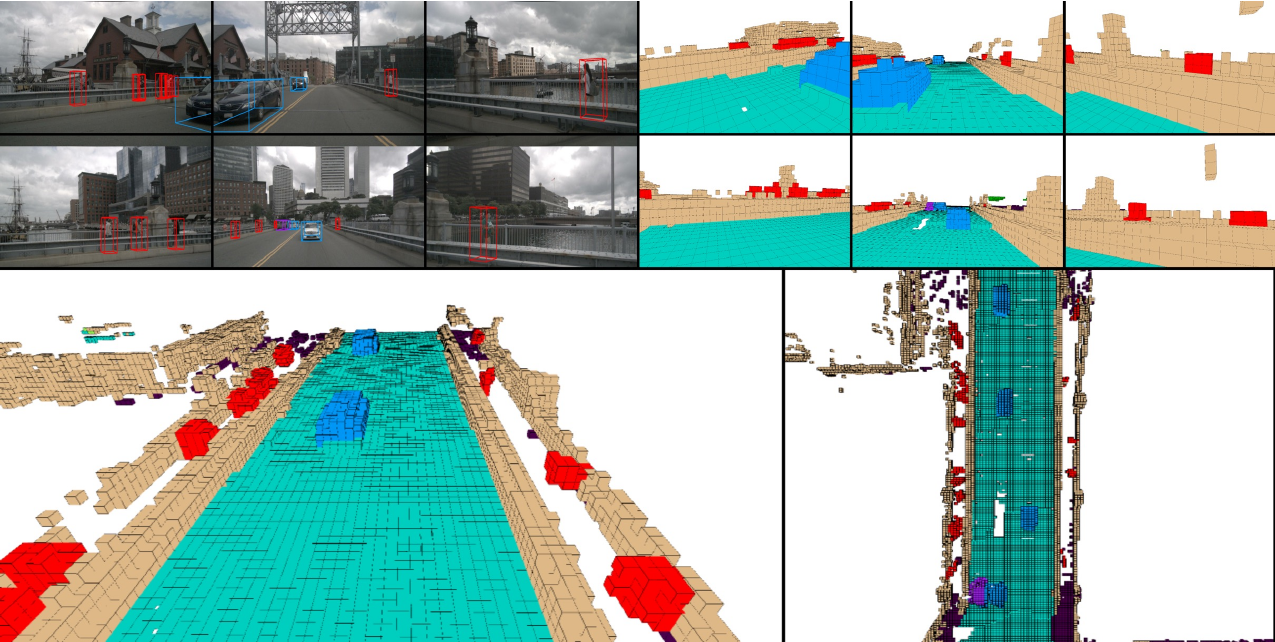}
  \vspace{-1em}
  \caption{\textbf{Visualization of 3D occupancy prediction on a bridge with moving vehicles and pedestrians.} The top-left shows object detection in the image space, the top-right presents occupancy prediction in the camera view, the bottom-left illustrates the top-front view, and the bottom-right depicts the top-down occupancy prediction results.}
  \vspace{-4em}
  \label{fig:visual_all_4}
\end{figure*}

\myparagraph{Additional qualitative results.}
Figures~\ref{fig:visual_all_1} to~\ref{fig:visual_all_4} present additional qualitative results in complex driving scenarios, further validating the robustness of the proposed framework.

\clearpage

\begin{figure*}[b!]
  \centering
    \includegraphics[width=.96\linewidth]{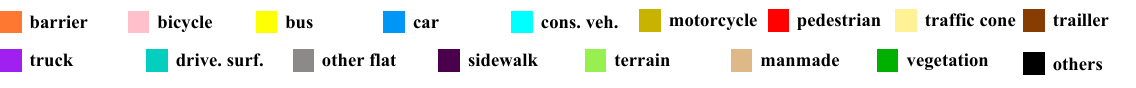}
  \vspace{-1em}
\end{figure*}

\begin{figure*}[b!]
  \centering
    \includegraphics[width=.96\linewidth]{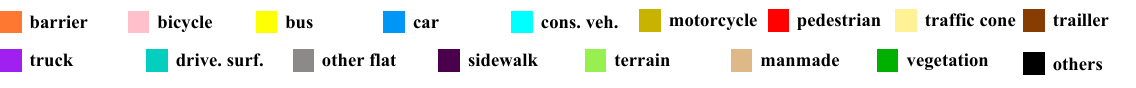}
  \vspace{-1em}
\end{figure*}

\clearpage

\bibliographystyle{splncs04}
\bibliography{main}
\end{document}